\documentclass[acmtog]{acmart}

\usepackage{subfigure}
\usepackage{graphicx}
\usepackage{amsmath}
\usepackage{amssymb}
\usepackage{algorithm}
\usepackage[noend]{algpseudocode}
\usepackage{color}
\usepackage{colortbl}
\usepackage{tabularx}

\usepackage{booktabs} % For formal tables
\newcommand{\cmt}[1]{}

\long\def\ignorethis#1{}

\newcommand{\etal}{{\em{et~al.}\ }}
\newcommand{\eg}{e.g.\ }
\newcommand{\ie}{i.e.\ }

\newcommand{\vc}[1]{\ensuremath{\mathbf{#1}}}

\newcommand{\argmax}{\operatornamewithlimits{argmax}}
\newcommand{\argmin}{\operatornamewithlimits{argmin}}

\newcommand{\pctab}{\hspace{0.2in}}

\acmPrice{15.00}

\setcopyright{acmlicensed}
\acmJournal{TOG}
\acmYear{2018}\acmVolume{37}\acmNumber{4}\acmArticle{144}\acmMonth{8} \acmDOI{10.1145/3197517.3201397}

\setcitestyle{square}

\begin{document}

%\acmSubmissionID{594}

% Title. 
\title{Learning Symmetric and Low-Energy Locomotion}
%\title{Learning Symmetric Low-Energy Locomotion}
%\title{Learning Locomotion Controller with Generic Locomotion Curriculum}

% Authors.

\author{Wenhao Yu}
\affiliation{%
  \department{School of Interactive Computing}
  \institution{Georgia Institute of Technology}}
\email{wyu68@gatech.edu}

\author{Greg Turk}
\affiliation{%
  \department{School of Interactive Computing}
  \institution{Georgia Institute of Technology}}
\email{turk@cc.gatech.edu}

\author{C.Karen  Liu}
\affiliation{%
  \department{School of Interactive Computing}
  \institution{Georgia Institute of Technology}}
\email{karenliu@cc.gatech.edu}

% This command defines the author string for running heads.
\renewcommand{\shortauthors}{Yu, Turk and Liu}

% abstract
\begin{abstract}
Learning locomotion skills is a challenging problem. To generate realistic and smooth locomotion, existing methods use motion capture, finite state machines or morphology-specific knowledge to guide the motion generation algorithms. Deep reinforcement learning (DRL) is a promising approach for the automatic creation of locomotion control. Indeed, a standard benchmark for DRL is to automatically create a running controller for a biped character from a simple reward function \cite{duan2016benchmarking}. Although several different DRL algorithms can successfully create a running controller, the resulting motions usually look nothing like a real runner. This paper takes a minimalist learning approach to the locomotion problem, without the use of motion examples, finite state machines, or morphology-specific knowledge. We introduce two modifications to the DRL approach that, when used together, produce locomotion behaviors that are symmetric, low-energy, and much closer to that of a real person. First, we introduce a new term to the loss function (not the reward function) that encourages symmetric actions. Second, we introduce a new curriculum learning method that provides modulated physical assistance to help the character with left/right balance and  forward movement. The algorithm automatically computes appropriate assistance to the character and gradually relaxes this assistance, so that eventually the character learns to move entirely without help. Because our method does not make use of motion capture data, it can be applied to a variety of character morphologies. We demonstrate locomotion controllers for the lower half of a biped, a full humanoid, a quadruped, and a hexapod. Our results show that learned policies are able to produce symmetric, low-energy gaits. In addition, speed-appropriate gait patterns emerge without any guidance from motion examples or contact planning.
\end{abstract}

% Learning locomotion skills is a challenging problem. To generate realistic and smooth locomotion, existing methods utilizes motion captures, finite state machines or morphology-specific knowledges to guide the motion generation algorithms. Directly learning a locomotion controller for arbitrary character morphology without prior knowledge about its motion usually suffers from difficulty in optimization and unnaturalness in the resulting motions. In this work, we propose two components that enable efficient learning of locomotion controllers using deep reinforcement learning for a large variety of character morphologies. The first component is an Assistance Controller (AC) providing external forces to the character, which improves the stability of learning, and the second component is a mirror symmetry loss that encourages learning symmetric control strategies. We then develop a learning curriculum to gradually reduce the strength of the Assistance Controller during training. We demonstrate learning of fully 3D locomotion controllers for different desired velocities on $4$ morphologies and show the emergence of symmetric and low-energy locomotion gaits learned entirely from scratch.


%keywords

% A "teaser" figure, centered below the title and authors and above the body of the work.
\begin{teaserfigure}
  \centering
  \includegraphics[width=6.0in]{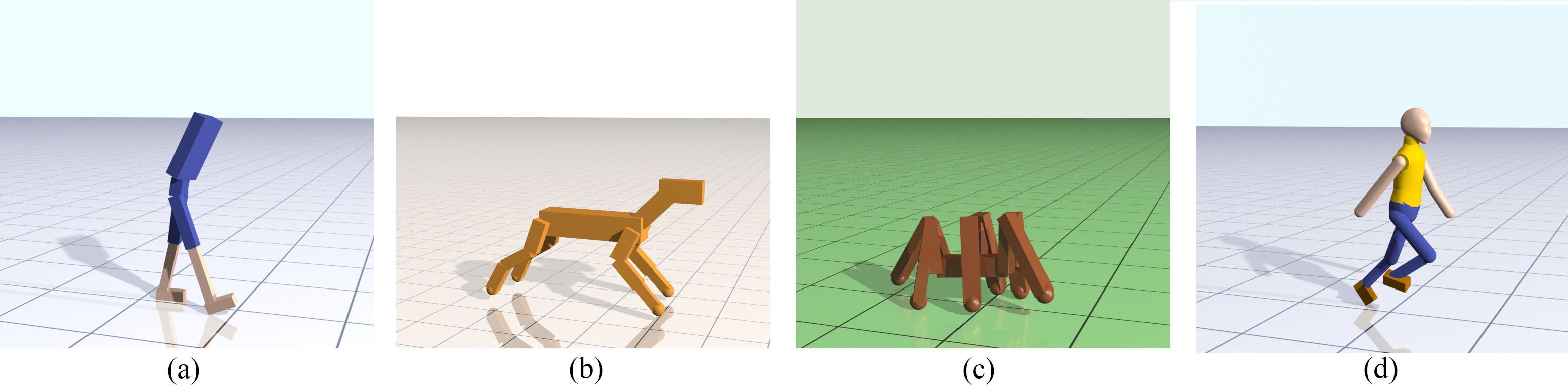}
  \caption{Locomotion Controller trained for different creatures. (a) Biped walking. (b) Quadruped galloping. (c) Hexapod Walking. (d) Humanoid running.}
  \label{fig:teaser}
\end{teaserfigure}

% Processes all of the front-end information and starts the body of the work.
\maketitle

\section{INTRODUCTION}

Creating an animated character that can walk is a fascinating challenge for graphics researchers and animators.  Knowledge from biomechanics, physics, robotics, and animation give us ideas for how to coordinate the virtual muscles of a character's body to move it forward while maintaining balance and style.  Whether physical or digital, the creators of these characters apply physics principles, borrow ideas from domain experts, use motion data, and undertake arduous trial-and-error to craft lifelike movements that mimic real-world animals and humans. While these characters can be engineered to exhibit locomotion behaviors, a more intriguing question is whether the characters can learn locomotion behaviors on their own, the way a human toddler can.

The recent disruptive development in Deep Reinforcement Learning (DRL) suggests that the answer is yes. Researchers indeed showed that artificial agents can learn some form of locomotion using advanced policy learning methods with a large amount of computation. Even though such agents are able to move from point A to point B without falling, the resulting motion usually exhibits jerky, high-frequency movements. The motion artifacts can be mitigated by introducing motion examples or special objectives in the reward function, but these remedies are somewhat unsatisfying as they sacrifice generality of locomotion principles and return partway to heavily engineered solutions.

This paper takes a minimalist approach to the problem of learning locomotion. Our hypothesis is that natural locomotion will emerge from simple and well-known principles in biomechanics, without the need of motion examples or morphology-specific considerations. If the agent can successfully learn in this minimal setting, we further hypothesize that our learning method can be generalized to agents with different morphologies and kinematic properties. Our approach is inspired by the recent work that applies reinforcement learning to train control policies represented as neural networks, but aims to address two common problems apparent in the motions produced by the existing methods. First, we observe that the motions appear much more energetic than biological systems. Second, an agent with perfectly symmetrical morphology often produces visibly asymmetrical motion, contradicting the observation in biomechanics literature that gaits are statistically symmetrical \cite{herzog1989asymmetries}. Therefore, we propose a new policy learning method that minimizes energy consumption and encourages gait symmetry to improve the naturalness of locomotion.

While most existing methods in motor skill learning penalize the use of energy in the reward function, the weighting of the penalty is usually relatively low for fear of negatively impacting the learning of the main task (\eg maintaining balance). As a result, the energy term serves merely as a regularizer and has negligible effect on preventing the agent from using excessive joint torques. We introduce a new curriculum learning method that allows for a high energy penalty while still being able to learn successfully using the existing policy learning algorithms. The curriculum provides modulated physical assistance appropriate to the current skill level of the learner, ensuring continuous progress toward successful locomotion with low energy consumption. Our algorithm automatically computes the assistive forces to help the character with lateral balance and forward movement. The curriculum gradually relaxes the assistance, so that eventually the character learns to move entirely without help.

In addition to energy consumption, gait symmetry offers both stable and adaptive locomotion that reduces the risk of falling \cite{patterson2008gait}. The symmetry of walking trajectories can be measured by established metric used in clinical settings \cite{nigg1987use}. However, directly using this metric in the reward function leads to two complications for policy gradient methods. First, the requirement of evaluating an entire trajectory introduces a delayed and sparse reward, resulting in a much harder learning problem. Second, the policy, especially at the beginning of learning, might not be able to produce structured, cyclic motion, rendering the symmetry metric ineffective in rewarding the desired behaviors. Our solution departs from the conventional metrics that measure the symmetry of the states. Instead, we measure the \emph{symmetry of actions} produced by the policy. We propose a mirror symmetry loss in the objective function to penalize the paired limbs for learning different control strategies.

Our evaluation shows that the agent can indeed learn locomotion that exhibits symmetry and speed-appropriate gait patterns and consumes relatively low-energy, without the need of motion examples, contact planning, and additional morphology-specific terms in the reward function. We further show that the same reward function with minimal change of weighting and the learning methodology can be applied to a variety of morphologies, such as bipeds, quadrupeds, or hexapods. We test our method against three baselines: learning without the mirror symmetry loss, learning without the curriculum, and learning without either component. The comparisons show that, without the curriculum learning, the trained policies fail to move forward or/and maintain balance. On the other hand, without the mirror symmetry loss, the learning process takes significantly more trials and results in asymmetric locomotion.

\section{RELATED WORK}

\subsection{Physically-based Character Control}

Obtaining natural human and animal motor skills has been a long-standing challenge for researchers in computer animation. Existing algorithms in character control usually require breaking the motion into more manageable parts or using motion data in order to generate natural-looking results \cite{Geijtenbeek2012}. One approach is the use of finite state machines (FSMs) \cite{Hodgins1995, Yin2007SSB, Jain2009, deLasa2010, Geijtenbeek2013, Wang2012, Coros2010, 2011TOGquadruped, felis2016synthesis,Wang2009OWC}. Yin \etal used FSMs to connect keyframes of the character poses, which is then combined with feedback rules to achieve balanced walking, skipping and running motion for humanoid characters \cite{Yin2007SSB}. A different application of FSMs can be seen in \cite{deLasa2010}, where they formulated the locomotion task as a Quadratic Programming (QP) problem and used an FSM model to switch between different objective terms to achieve walking motion. Although this class of techniques can successfully generate plausible character motions, it is usually difficult to generalize them to non-biped morphologies or arbitrary tasks.

An alternative approach to obtain natural character motions is to incorporate motion data such as videos \cite{Wampler2014} or motion capture data \cite{Lee2014LCM, daSilva2008ISS, Ye2010OFC, Muico2009CNC, Sok2007SBB,Liu2005LPB, lee2010data}. Despite the high fidelity motion this approach can generate, the requirement of motion data does not allow its application to arbitrary character morphologies as well as generalization to novel control tasks.

Apart from designing finite state machines or using motion data, reward engineering combined with trajectory optimization has also been frequently applied to generate physically-based character motions \cite{al2013trajectory,Wampler2009,Mordatch2013AHL, Mordatch2012DCB,HA2014ITD}. Al Borno \etal demonstrated a variety of humanoid motor skills by breaking a sequence of motion into shorter windows, and for each window a task-specific objective is optimized \cite{al2013trajectory}. Mordatch \etal applied the Contact-Invariant-Optimization algorithm to generate full-body humanoid locomotion with muscle-based lower-body actuations \cite{Mordatch2013AHL}. Symmetry and periodicity of the motion was explicitly enforced to generate realistic locomotion. Trajectory-optimization based methods provide a general framework for synthesizing character motions. However, a few common drawbacks for this category of methods include: they are usually off-line methods, they are sensitive to large perturbations, and they require explicitly modeling of system dynamics. To overcome the first two issues, Mordatch \etal trained a neural network policy using data generated from a trajectory optimization algorithm and demonstrated interactive control of character locomotion with different morphologies \cite{mordatch2015interactive}. In this work, we aim to develop an algorithm that can synthesize plausible locomotion controllers with minimal prior knowledge about the character, the target motion, and the system dynamics.

\subsection{Reinforcement Learning}

Reinforcement Learning (RL) provides a general framework for modeling and solving control of physics-based characters. In computer animation, policy search methods, a sub-area of RL, have been successfully applied in optimizing control policies for complex motor skills, such as cartwheels \cite{Liu2016GLC}, monkey-vaults \cite{HA2014ITD}, bicycle stunts \cite{Tan2014LBS}, skateboarding \cite{liu2017learning} and locomotion \cite{Peng2015DTT,Peng2016TLS, Peng2017DDL, 2014cgps,Geijtenbeek2013, won2017train}. Two major classes of policy search methods used in the previous work include sampling-based methods \cite{Tan2014LBS,Geijtenbeek2013} and gradient-based methods \cite{Liu2016GLC,Peng2016TLS, Peng2017DDL, 2014cgps}. %\karen{I think we are abusing the term RL a bit here. I would not call Jie's swimming paper an RL approach. If we want to include any papers that use sample-based method to optimize motion trajectory and then pd-track the trajectory, then Libin's early work and Pertu's 2015 siggraph are also in this category. Similarly, Jack's 2009 paper seems a stretched to be called RL. All it does is to run CMA on Simbicon's parameters. On the other hand, another paper that should've been cited is Terrain-adaptive bipedal locomotion control by Jia-chi Wu.}

A representative example of sampling-based method is CMA-ES, which iterates between evaluating a set of sampled parameters and improving the sampling distribution \cite{hansen1996adapting}. This class of policy search methods is relatively robust to local minima and does not require gradient information. However, the sample number and memory requirement usually scales with the number of model parameters, making it less suitable to optimize models with many parameters such as deep neural networks. To combat this constraint on policy complexity, researchers resort to specialized controller design and a fine-tuned reward structure.

On the other hand, gradient-based methods naturally fit into the stochastic gradient descent framework, an algorithm that has been successfully demonstrated to train deep neural networks with hundreds of millions of parameters \cite{lecun2015deep}. Gradient-based methods like REINFORCE exhibit large variance in the estimated gradient, limiting their application to relatively simple control problems \cite{sutton2000policy}. Recent developments in deep reinforcement learning have seen significant progress on improving the gradient estimation accuracy and stability of training, leading to algorithms such as DDPG \cite{lillicrap2015continuous}, A3C \cite{mnih2016asynchronous}, TRPO \cite{schulman2015trust}, and PPO \cite{schulman2017proximal}, which can be used to solve complex motor control problems. For example, by combining TRPO with Generalized Advantage Estimation \cite{schulman2015high}, Schulman \etal demonstrated learning of locomotion controllers for a 3D humanoid character. Later, they proposed Proximal Policy Optimization (PPO), which further improved the data efficiency of the algorithm \cite{schulman2017proximal}. Despite the impressive feat of tackling the 3D biped locomotion problem from scratch, the resulting motion usually looks jerky and unnatural. Peng \etal achieved significantly more natural-looking biped locomotion by combining motion capture data with an actor-critic algorithm \cite{Peng2017DDL}.

\subsection{Curriculum Learning}

Our approach is also inspired by works in curriculum learning (CL). The general idea behind CL is to present the training data to the learning algorithm in an order of increasing complexity \cite{bengio2009curriculum}. Researchers in machine learning have shown that CL can improve performance and efficiency of learning problems such as question-answering \cite{graves2017automated}, classification \cite{bengio2009curriculum, pentina2015curriculum}, navigation \cite{matiisen2017teacher,held2017automatic} and game playing \cite{narvekar2016source}.

Curriculum learning has also been applied to learning motor skills in character animation and robotics. Florensa \etal demonstrated successful training of robot manipulation and navigations controllers, where the policies are trained with initial states that are increasingly farther from a goal state \cite{florensa2017reverse}. Pinto \etal applied CL to improve the efficiency of learning of a robot grasping controller \cite{pinto2016supersizing}. Karpathy et al trained a single legged character to perform acrobatic motion by decomposing the motion into a high-level curriculum with a few sub-tasks and learned each sub-task with a low-level curriculum\cite{karpathy2012curriculum}. In learning locomotion controllers, van de Panne \etal applied external torques to the character in order to keep it upright during training and demonstrated improved learning performance \cite{van1995guided}. Similarly, Wu \etal used helper forces to assist in optimizing a set of locomotion controllers, which are then used by a foot-step planner to controller the character to walk on uneven terrains \cite{wu2010terrain}. Our method shares the similar idea of using an external controller to assist the character in learning locomotion gaits, but differs from their works in two key aspects. First, in addition to balance assistance, our curriculum learning also provides \textit{propelling assistance} which shows significant improvement over balance assistance alone. Second, our method gradually reduces the strength of the assistance forces without affecting the reward function, while their method explicitly minimizes the assistances force in the optimization which might require objective function tuning. Yin \etal applied continuation method to search for a curriculum path that gradually adapt a nominal locomotion controller to different tasks such as stepping over obstacle or walking on ice \cite{yin2008continuation}. More recently, Heess \etal demonstrated learning of agile locomotion controllers in a complex environment using deep Reinforcement Learning \cite{heess2017emergence}. They applied PPO to train the agents on environments with increasing complexity, which provided a environment-centered curriculum. Similarly, we also find that efficiency of curriculum learning can be improved by using a environment-centered curriculum.

% I don't think this is relevant to the discussion of similarity between our work and van de Panne's:
% and 3) we propose a Mirror Symmetry Loss to further improve the symmetry of the motion. 

\section{Background: Policy Learning}
\label{sec:policy_learning}

The locomotion learning process can be modeled as a Markov Decision Process (MDP), defined by a tuple: $(\mathcal{S}, \mathcal{A}, r, \rho_0, P, \gamma)$, where $\mathcal{S}$ is the state space, $\mathcal{A}$ is the action space, $r: \mathcal{S} \times \mathcal{A} \mapsto \mathbb{R}$ is the reward function, $\rho_0$ is the initial state distribution, $P: \mathcal{S} \times \mathcal{A} \mapsto \mathcal{S}$ is the transition function and $\gamma$ is the discount factor. Our goal is to solve for the optimal parameters $\theta$ of a policy $\pi_\theta: \mathcal{S} \times \mathcal{A} \mapsto \mathbb{R}$, which maximizes the expected long-term reward:
\begin{equation}
\pi_{{\theta}^*} = \argmax_\theta \;\;\mathbb{E}_{\vc{s}{\sim}\rho_0}[V^\pi(\vc{s})],
\end{equation}
where the value function of a policy, $V^\pi: \mathcal{S} \mapsto \mathbb{R}$, is defined as the expected long-term reward of following the policy $\pi_\theta$ from some input state $\vc{s}_t$: 
\begin{equation} 
V^\pi(\vc{s}_t) = \mathbb{E}_{\vc{a}_t \sim \pi {,}\vc{s}_{t+1} \sim P{,}...}[\sum_{i=0}^\infty \gamma^i r(\vc{s}_{t+i}, \vc{a}_{t+i})].
\end{equation}

\subsection{Learning Locomotion Policy}
\label{sec:learning_locomotion}
In the context of locomotion learning problem, we define a state as $\vc{s} = [\vc{q}, \dot{\vc{q}}, \vc{c}, \hat{v}]$, where $\vc{q}$ and $\dot{\vc{q}}$ are the joint positions and joint velocities. $\vc{c}$ is a binary vector with the size equal to the number of end-effectors, indicating the contact state of the end-effectors ($1$ in contact with the ground and $0$ otherwise). $\hat{v}$ is the target velocity of the center of mass in the forward direction. The action $\vc{a}$ is simply the joint torques generated by the actuators of the character.

Designing a reward function is one of the most important tasks in solving a MDP. In this work, we use a generic reward function for locomotion similar to those used in RL benchmarks \cite{duan2016benchmarking} \cite{DBLP:journals/corr/BrockmanCPSSTZ16} \cite{openai}. It consists of three objectives: move forward, balance, and use minimal actuation.
\begin{equation}
r(\vc{s},\vc{a}) =  w_v E_{v}(\vc{s}) + E_u(\vc{s}) + w_l E_l(\vc{s}) + E_a + w_e E_{e}(\vc{a}).
\end{equation}

The first term of the reward function, $E_v = -|\bar{v}(\vc{s}) - \hat{v}|$, encourages the character to move at the desired velocity $\hat{v}$. $\bar{v}(\vc{s})$ denotes the average velocity in the most recent $2$ seconds. The next three terms are designed to maintain balance. $E_u = - (w_{u_x}|\phi_x(\vc{s})| + w_{u_y}|\phi_y(\vc{s})| + w_{u_z}|\phi_z(\vc{s})|)$ rewards the character for maintaining its torso or head upright, where $\phi(\vc{s})$ denotes the orientation of the torso or head. $E_l = -|c_z(\vc{s})|$ penalizes deviation from the forward direction, where $c_z(\vc{s})$ computes the center of mass (COM) of the character in the frontal axis. $E_a$ is the alive bonus which rewards the character for not being terminated at the current moment. A rollout is terminated when the character fails to keep its COM elevated along the forward direction, or to keep its global orientation upright. Finally, $E_e = -\|\vc{a}\|$ penalizes excessive joint torques, ensuring minimal use of energy. Details on the hyper-parameters related to the reward function and the termination conditions are discussed in Section \ref{sec:results}.

\subsection{Policy Gradient Algorithm}
Policy gradient methods have demonstrated success in solving such a high-dimensional, continuous MDP. In this work, 
we use Proximal Policy Optimization (PPO) \cite{schulman2017proximal} to learn the optimal locomotion policy because it provides better data efficiency and learning performance than the alternative learning algorithms. Our method can also be easily applied to other learning algorithms such as Trust Region Policy Optimization (TRPO) \cite{schulman2015trust} or Deep Deterministic Policy Gradient (DDPG) \cite{lillicrap2015continuous}.

Like many policy gradient methods, PPO defines an advantage function as $A^\pi(\vc{s}_t, \vc{a}) = Q^\pi(\vc{s}, \vc{a}) - V^\pi(\vc{s})$, where $Q^\pi$ is the state-action value function that evaluates the return of taking action $\vc{a}$ at state $\vc{s}$ and following the policy $\pi_\theta$ thereafter. However, PPO minimizes a modified objective function to the original MDP problem:
\begin{align}
L_{PPO}({\theta}) = -\mathbb{E}_{\vc{s}_0, \vc{a}_0, \vc{s}_1, ...} [\min (&r(\theta)A_t, clip(r(\theta), 1-\epsilon, 1+\epsilon)A_t)],
\end{align}
where $r(\theta) = \frac{\pi_\theta(\vc{a}|\vc{s})}{\pi_{\theta_{old}}(\vc{a}|\vc{s})}$ is the importance re-sampling term that enables us to use data sampled under an old policy $\pi_{\theta_{old}}$ to estimate expectation for the current policy $\pi_{\theta}$. The $min$ and the $clip$ operators together ensure that $\pi_{\theta}$ does not change too much from $\pi_{\theta_{old}}$. More details in deriving the objective functions can be found in the original paper on PPO \cite{schulman2017proximal}.
%Incorporating the Mirror Symmetry Loss, the final optimization problem for learning the locomotion policy can be defined as:

%\begin{equation}
%\pi_{\hat{\theta}} = \arg \min_\theta (L_{PPO}(\theta) + \omega_{mirror} L_{mirror}(\theta)),
%\end{equation}
%where $\omega_{mirror}$ is the weight of the Mirror Symmetry Loss. We use $\omega_{mirror} = 4$ for all the examples.

%The value function measures the return of a trajectory starting from $\vc{s}_0$ and follows the policy $\pi_\theta$. 
%We also define an state-action value function that 
%\begin{equation}
%Q(\vc{s}_0, \vc{a}) = R(\vc{s}_0, \vc{a}) + \gamma \mathbb{E}_{\vc{s}_1{\sim}T(\vc{s}_0, \vc{a})}[V(\vc{s}_1)],
%\end{equation}
%which estimates the return of a trajectory starting with taking action $\vc{a}$ at state $\vc{s}_0$ and follows the policy $\pi_\theta$ afterwards. 

\section{Locomotion Curriculum Learning}
Learning locomotion directly from the principles of minimal energy and gait symmetry is difficult without additional guidance from motion examples or reward function shaping. Indeed, a successful locomotion policy must learn a variety of tasks often with contradictory goals, such as maintaining balance while propelling the body forward, or accelerating the body while conserving energy. One approach to learning such a complex motor skill is to design a curriculum that exposes the learner to a series of tasks with increasing difficulty, eventually leading to the original task. 

Our locomotion curriculum learning is inspired by physical learning aids that provide external forces to simplify the motor tasks, such as exoskeletons for gait rehabilitation or training wheels for riding a bicycle. These learning aids create a curriculum to ease the learning process and will be removed when the learner is sufficiently skilled at the original task. To formalize this idea, we view the curriculum as a continuous Euclidean space parameterized by curriculum variables $\vc{x} \in \mathbb{R}^n$. The learning begins with the simplest lesson $\vc{x}_0$ for the beginner learner, gradually increasing the difficulty toward the original task, which is represented as the origin of the curriculum space (\ie $\vc{x} = \vc{0}$). With this notion of a continuous curriculum space, we can then develop a continuous learning method by finding the optimal path from $\vc{x}_0$ to the origin in the curriculum space.

%\begin{figure}[ht]
%  \centering
%  \includegraphics[width=\linewidth]{images/overview}
%  \caption{Overview of our method. Black arrow denotes simulation steps, red arrow denotes information flow and blue arrow denotes learning steps.}
%  \label{fig:overview}
%\end{figure}

Similar to the standard policy gradient method, at each learning iteration, we generate rollouts from the current policy, use the rollout to estimate the gradients of the objective function of policy optimization, and update the policy parameters $\theta$ based on the gradient. With curriculum learning, we introduce a virtual assistant to provide assistive forces to the learner during rollout generation. The virtual assistant is updated at each learning iteration such that it provides assistive forces appropriate to the current skill level of the learner.

Two questions remain in our locomotion curriculum learning algorithm. First, what is the most compact set of parameters for the virtual assistant such that locomotion skills can be effectively learned through curriculum? Second, what is the appropriate curriculum schedule, \ie how much assistive force should we give to the learner at each moment of learning?

\subsection{Virtual Assistant}
Our virtual assistant provides assistive forces to simplify the two main tasks of locomotion: moving forward and maintaining lateral balance. The lateral balancing force is applied along the frontal axis (left-right) of the learner, preventing it from falling sideway. The propelling force is applied along the sagittal axis, pushing the learner forward to reach the desired velocity. With these two assistive forces, the learner can focus on learning to balance in the sagittal plane as well as keeping the energy consumption low. 

Both lateral balancing force and propelling force are produced by a virtual proportional-derivative (PD) controller placed at the pelvis of the learner, as if an invisible spring is attached to the learner to provide support during locomotion. Specifically, the PD controller controls the lateral position and the forward velocity of the learner. The target lateral position is set to $0$ for maintaining lateral balance while the target forward velocity is set to the desired velocity $\bar{v}$ for assisting the learner moving forward. 

Different levels of assistance from the virtual assistant create different lesson for the learner. We use the stiffness coefficient $k_p$ and the damping coefficient $k_d$ to modulate the strength of the balancing and propelling forces respectively. As such, our curriculum space is parameterized by $\vc{x} = (k_p, k_d)$. Any path from $\vc{x}_0$ to $(0, 0)$ constitutes a curriculum for the learner.

Our implementation of the virtual assistant is based on the stable proportional-derivative (SPD) controller proposed by Tan \etal \shortcite{tan2011stable}. The SPD controller provides  a few advantages. First, it does not require any pre-training and can be applied to any character morphology with little tuning. In addition, it provides a smooth assistance in the state space, which facilitates learning. Finally, it is unconditionally stable, allowing us to use large controller gains without introducing instability.  

\subsection{Curriculum Scheduling}

\begin{figure}[ht]
  \centering
  \includegraphics[width=\linewidth]{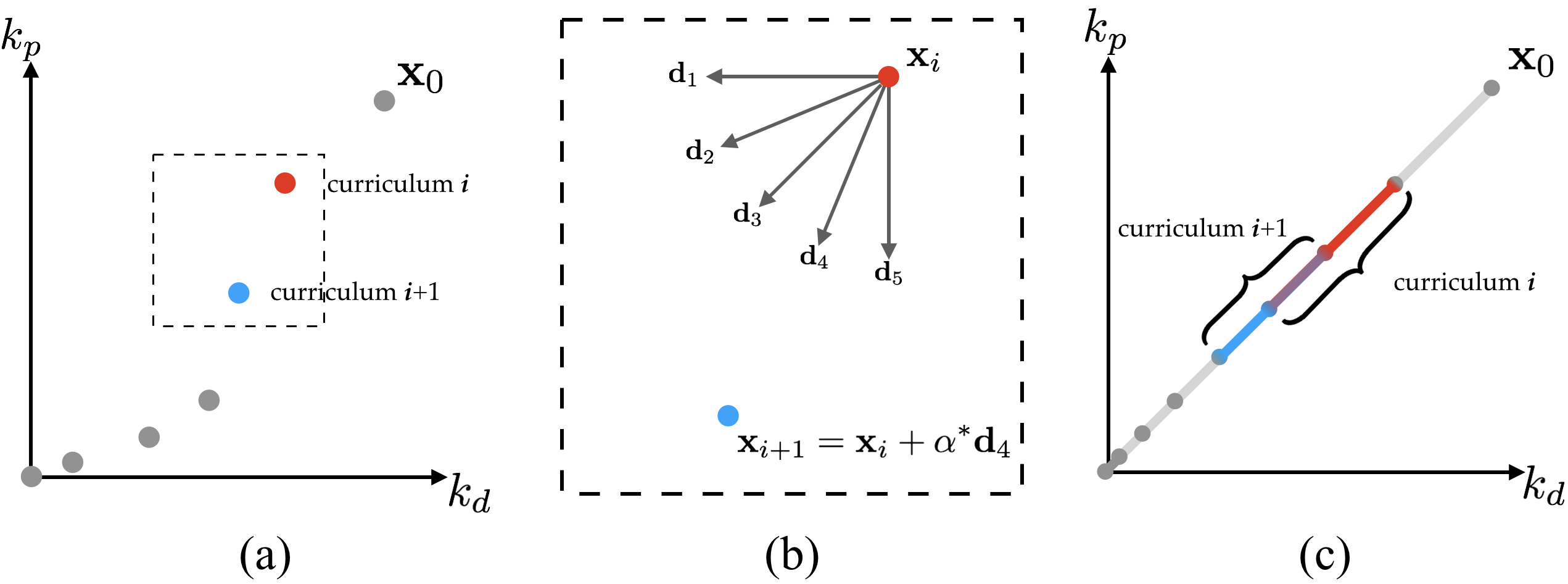}
  \vspace{-3mm}
  \caption{(a) The learner-centered curriculum determines the lessons adaptively based on the current skill level of the agent, resulting in a piece-wise linear path from $\vc{x}_0$ to the origin. (b) In each learning iteration, the learner-centered curriculum finds the next point in the curriculum space using a simple search algorithm that conducts 1D line-searches along five direction. The goal is to find the largest step size, $\alpha^*$, such that the current policy can still reach $60\%$ of the original return $\bar{R}$. (c) Environment-centered curriculum follows a series of predefined lessons along a linear path from $\vc{x}_0$ to the origin. It introduces the learner to a range of lessons in one curriculum learning iteration, resulting in a set of co-linear, overlapping line segments.}
  \label{fig:curr_schedule}
\end{figure}

The goal of curriculum scheduling is to systematically and gradually reduce the assistance from the initial lesson $\vc{x}_0$ and ultimately achieve the final lesson in which the assistive force is completely removed. Designing such a schedule is challenging because an aggressive curriculum that reduces the assistive forces too quickly can fail the learning objectives while a conservative curriculum can lead to inefficient learning.

We propose two approaches to the problem of curriculum scheduling (Figure \ref{fig:curr_schedule}): Learner-centered curriculum and Environment-centered curriculum. The learner-centered curriculum allows the learner to decide the next lesson in the curriculum space, resulting in a piece-wise linear path from $\vc{x}_0$ to the origin. The environment-centered curriculum, on the other hand, follows a series of predefined lessons. However, instead of focusing on one lesson at a time, it exposes the learner to a range of lessons in one curriculum learning iteration, resulting in a set of co-linear, overlapping line segments from $\vc{x}_0$ to the origin of the curriculum space.

\subsubsection{Learner-centered curriculum}
The learner-centered curriculum determines the lessons adaptively based on the current skill level of the agent (Algorithm \ref{alg:learner_center_curriculum}). We assume that the initial lesson $\vc{x}_0$ is sufficiently simple such that the standard policy learning algorithm can produce a successful policy $\pi$, which generates rollouts $\mathcal{B}$ with average return, $\bar{R}$. We then update the lesson to make it more challenging (Algorithm \ref{alg:update_curriculum}) and proceed to the main learning loop. At each curriculum learning iteration, we first update $\pi$ by running one iteration of the standard policy learning algorithm. If the average of the rollouts from the updated policy can reach $h\%$ of the original return $\bar{R}$ ($h = 80$), we update the lesson again using Algorithm \ref{alg:update_curriculum}. The curriculum learning loop terminates when the magnitude of $\vc{x}$ is smaller than $\epsilon$ ($\epsilon = 5$). We run a final policy learning without any assistance, \ie $\vc{x} = (0, 0)$. At this point, the policy learns this final lesson very quickly.

Given the current lesson $\vc{x}_i$, the goal of Algorithm \ref{alg:update_curriculum} is to find the next point in the curriculum space that is the closest to the origin while the current policy can still retain some level of proficiency. Since it is only a two-dimensional optimization problem and the solution $\vc{x}_{i+1}$ lies in $\|\vc{x}_{i+1}\| - \|\vc{x}_i\| < 0$ and is strictly positive component-wise, we implement a simple search algorithm that conducts 1D line-searches along five directions from $\vc{x}_i$: (-1, 0), (-1, -0.5), (-1, -1), (-0.5, -1), (0, -1). For each direction $\vc{d}$, the line-search will return the largest step size, $\alpha$, such that the current policy can still reach $l\%$ of the original return $\bar{R}$ ($l = 60$) under the assistance $\vc{x}_i + \alpha \vc{d}$. Among five line-search results, we choose the largest step size, $\alpha^*$ along the direction $\vc{d}^*$ to create the next lesson: $\vc{x}_{i+1} = \vc{x}_i + \alpha^* \vc{d}^*$.

\subsubsection{Environment-centered curriculum}
Instead of searching for the next lesson, the environment-centered curriculum updates the lessons along a predefined linear path from $\vc{x}_0$ to the origin (Algorithm \ref{alg:env_cent_cur}). In addition, the learner is trained with a range of lessons $[\vc{x}_{begin}, \vc{x}_{end}]$ in each curriculum learning iteration. Specifically, the learner will be exposed to an environment in which the virtual assistant starts with $\vc{x}_{begin}$ and reduces its strength gradually to $\vc{x}_{end}$ at the end of each rollout horizon. The formula of strength reduction from $\vc{x}_{begin}$ to $\vc{x}_{end}$ can be designed in several ways. We use a simple step function to drop the strength of the virtual assistant by $k\%$ every $p$ seconds ($k = 25$ and $p = 3$). Each step in the step function can be considered a learning milestone. Training with a range of milestones in a single rollout prevents the policy from overfitting to a particular virtual assistant, leading to more efficient learning.

In each learning iteration, we first run the standard policy learning for one iteration, with the environment programmed to present the current range of lessons $[\vc{x}_{begin}, \vc{x}_{end}]$. After the policy is updated, we evaluate the performance of the policy using two conditions. If the policy meets both conditions, we update the range of lessons to $k \% \cdot [\vc{x}_{begin}, \vc{x}_{end}]$. Note that the updated $\vc{x}_{begin}$ is equivalent to the second milestone of the previous learning iteration, resulting in some overlapping lessons in two consecutive curriculum learning iterations (See Figure \ref{fig:curr_schedule}c). The overlapping lessons are an important aspect of the environment-centered curriculum learning because they allow the character to bootstrap its current skill when learning a new set of predefined lessons. Similar to the learner-centered curriculum, the curriculum learning loop terminates when the magnitude of $\vc{x}_{begin}$ is smaller than $\epsilon$ ($\epsilon = 5$), and we run a final policy learning without any assistance, \ie $\vc{x}_{begin} = (0, 0)$ and $\vc{x}_{end} = (0, 0)$.

The first condition for assessing the progress of learning checks whether the learner is able to reach the second milestone in each rollout. That is, the agent must stay balance for at least $2p$ seconds. Using this condition alone, however, might result in a policy that simply learns to stand still or move minimally in balance. Therefore, we use another condition that requires the average return of the policy to reach a pre-determined threshold, $\bar{R}$, which is $g\%$ of the return from the initial policy trained with full assistance ($g = 70$).

\begin{algorithm}[t]
\caption{Learner-Centered Curriculum Learning}\label{alg:learner_center_curriculum}
\begin{algorithmic}[1]
\State $\vc{x} = \vc{x}_0$ \;
\State $[\pi, \mathcal{B}] \leftarrow$ PolicyLearning($\vc{x}$) \;
\State $\bar{R} \leftarrow$ AvgReturn($\mathcal{B}$) \;
\State $\vc{x} \leftarrow$ UpdateLesson($\vc{x}$, $\bar{R}$, $\pi$) \;
\While {$||\vc{x}|| \geq \epsilon$ }
\State $[\pi, \mathcal{B}] \leftarrow$ OneIterPolicyLearning($\vc{x}$) \;
\State $R$ $\leftarrow$ AvgReturn($\mathcal{B}$) \;
\If {$R \geq h\% \cdot \bar{R}$}
\State $\vc{x} \leftarrow$ UpdateLesson($\vc{x}$, $\bar{R}$, $\pi$) \;
\EndIf
\EndWhile
\State $[\pi, \mathcal{B}] \leftarrow$ PolicyLearning($(0,0)$) \;
\Statex
\Return $\pi$
\end{algorithmic}
\end{algorithm}

\begin{algorithm}[t]
\caption{Update Lesson}\label{alg:update_curriculum}
\begin{algorithmic}[1]
%\Function{UpdateCurriculum}{$\vc{x}, \bar{R}, \mathcal{B}$}
\Statex \textbf{input:} $\vc{x}, \bar{R}, \pi$
\State $\mathcal{D} = [(-1,0), (-1,-0.5), (-1,-1), (-0.5, -1), (0, -1)]$ \;
\State $\vc{x}_{min} = (\infty, \infty)$ \;
\For{each $\vc{d} \in \mathcal{D}$}
\State $\alpha^*=\argmax_\alpha \;\alpha$ \;
\Statex \ \ \ \ \ \ \ \ \ \ \  s.t. EvalReturn($\vc{x}+ \alpha\vc{d}, \pi) > l \% \cdot \bar{R}$ \;
\If{$||\vc{x} + \alpha^*\vc{d}|| < ||\vc{x}_{min}||$}
\State $\vc{x}_{min} = \vc{x} + \alpha^*\vc{d}$
\EndIf
\EndFor
\Return $\vc{x}_{min}$
%\EndFunction
\end{algorithmic}
\end{algorithm}

\begin{algorithm}[t]
\caption{Environment-Centered Curriculum Learning}\label{alg:env_cent_cur}
\begin{algorithmic}[1]
\State $\vc{x}_{begin} = \vc{x}_0$ \;
\State $\vc{x}_{end} = (k^2)\% \cdot \vc{x}_{begin}  $ \;
\State $[\pi, \mathcal{B}] \leftarrow$ PolicyLearning($\vc{x}_{begin},\vc{x}_{end}$) \;
\State $\bar{R} \leftarrow$ $g \% \cdot$ AvgReturn($\mathcal{B}$) \;
%\While {$\vc{x}_{begin} \neq (0, 0)$ }
\While {$||\vc{x}_{begin}|| \geq \epsilon$ }
\State $[\pi, \mathcal{B}] \leftarrow$ OneIterPolicyLearning($\vc{x}_{begin},\vc{x}_{end}$) \;
\If {BalanceTest($\mathcal{B}$) and AvgReturn$(\mathcal{B}) > \bar{R}$}
\State $\vc{x}_{begin} = k \% \cdot \vc{x}_{begin}$ \;
\State $\vc{x}_{end} = k \% \cdot \vc{x}_{end}$ \;
%\If{$||\vc{x}_{begin}|| < 5$}
%\Sate $\vc{x}_{begin} = \vc{x}_{end} = (0,0)$ \;
%\EndIf
\EndIf
\EndWhile
\State $[\pi, \mathcal{B}] \leftarrow$ PolicyLearning($(0,0),(0,0)$) \;
\Statex
\Return $\pi$ 
\end{algorithmic}
\end{algorithm}

%\begin{algorithm}[t]
%\caption{Environment-Centered Curriculum Learning}\label{alg:env_cent_cur}
%\begin{algorithmic}[1]
%\State $\vc{x}_{begin} = \vc{x}_0$ \;
%\State $\vc{x}_{end} = p^2\vc{x}_{begin}  $ \;
%\State $T = -\infty$ \;
%\While {$\vc{x}_{begin} \neq (0, 0)$ }
%\State $[\pi, \mathcal{B}] \leftarrow$ OneIterPolicyLearning($\vc{x}_{begin},\vc{x}_{end}$) \;
%\If {BalanceTest($\mathcal{B}$) and AvgReturn$(\mathcal{B}) > T$}
%\State $\vc{x}_{begin} = p\vc{x}_{begin}$ \;
%\State $\vc{x}_{end} = p\vc{x}_{end}$ \;
%\If{$||\vc{x}_{begin}|| < 5$}
%\State $\vc{x}_{begin} = \vc{x}_{end} = (0,0)$ \;
%\EndIf
%\If{$T = -\infty$}
%\State T = AvgReturn$(\mathcal{B})$ \;
%\EndIf
%\EndIf
%\EndWhile
%\end{algorithmic}
%\end{algorithm}

\section{Mirror Symmetry Loss}
\label{sec:mirror_sym_loss}
Symmetry is another important characteristic of a healthy gait. Assessing gait symmetry usually requires at least an observation of a full gait cycle. This requirement poses a challenge to policy learning because the reward cannot be calculated before the end of the gait cycle, leading to a delayed reward function. We propose a new way to encourage gait symmetry by measuring the symmetry of \emph{actions} instead of \emph{states}, avoiding the potential issue of delayed reward.

Imaging a person who is standing in front of a floor mirror with her left hand behind her back. If she uses the right hand to reach for her hat, what we see in the mirror is a person with her right hand behind her back reaching for a hat using her left hand. Indeed, if the character has a symmetric morphology, the action it takes in some pose during locomotion should be the mirrored version of the action taken when the character is in the mirrored pose. This property can be expressed as:
\begin{equation}
\pi_{\theta}(\vc{s}) = \Psi_a(\pi_{\theta}(\Psi_o(\vc{s}))),
\label{eqn:mirror}
\end{equation}
where $\Psi_a(\cdot)$ and $\Psi_o(\cdot)$ maps actions and states to their mirrored versions respectively. We overload the  notation $\pi_\theta$ to represent the mean action of the stochastic policy. Enforcing Equation \ref{eqn:mirror} as a hard constraint is difficult for standard policy gradient algorithms, but we can formulate a soft constraint and include it in the objective function for policy optimization:
\begin{equation}
L_{sym}(\theta) = \sum_{i=0}^B||\pi_{\theta}(\vc{s}_i) - \Psi_a(\pi_{\theta}(\Psi_o(\vc{s}_i)))||^2,
\label{eqn:mirror-loss}
\end{equation}
where $B$ is the number of simulation samples per iteration. We use $20,000$ samples in all of our examples. Since Equation \ref{eqn:mirror-loss} is differentiable with respect to the policy parameters $\theta$, it can be combined with the standard reinforcement learning objective and optimized using any gradient-based RL algorithm.

%One may also include the Mirror Symmetry Loss in the reward function instead of the objective function. However, we note that doing this may not be as effective in the PPO framework. In particular, PPO makes multiple updates on the policy in between rollout collection sessions. Including the Mirror Symmetry Loss in the objective function directly would guarantee that the loss is always computed using the latest policy parameters $\theta$, while the reward value would not be changed after a rollout is generated.

Incorporating the mirror symmetry loss, the final optimization problem for learning the locomotion policy can be defined as:
\begin{equation}
\pi_{{\theta}^*} = \argmin_\theta \;\;L_{PPO}(\theta) + w L_{sym}(\theta),
\end{equation}
where $w$ is the weight to balance the importance of the gait symmetry and the expected return of the policy ($w = 4$). Note that the mirror symmetry loss is included in the objective function of the policy optimization, rather than in the reward function of the MDP. This is because $L_{sym}$ explicitly depends on the policy parameters $\theta$, thus adding $L_{sym}$ in the reward function would break the assumption in the Policy Gradient Theorem \cite{sutton2000policy}. That is, changing $\theta$ should change the probability of a rollout, not its return. If we included $L_{sym}$ in the reward function, it would change the return of the rollout when $\theta$ is changed. Instead, we include $L_{sym}$ in the objective function and calculate its gradient separately from that of the $L_{PPO}$, which depends on the Policy Gradient Theorem.

%Ideally, the policy learning algorithm should search only in the space of policy with such property for these problems. However, enforcing this analytically is non-trivial as typical controllers treat different parts of the character independently. As a result, the trained controller tends to use asymmetric control strategies and generates unnatural motion. To encourage the algorithm in searching policies with mirror symmetry property, we formulate it as a soft constraint:

%\input{overview}
%\input{method}
\section{RESULTS}
\label{sec:results}

We evaluate our method on four characters with different morphologies and degrees of freedom. The input to the control policy includes $\vc{s} = [\vc{q}, \dot{\vc{q}}, \vc{c}, \hat{v}]$ and the output is the torque generated at each actuated joint, as described in Section \ref{sec:learning_locomotion}. Because the character is expected to start at zero velocity and accelerate to the target velocity, the policy needs to be trained with a range of velocities from zero to the target. We include $\hat{v}$ in the input of the policy to modulate the initial acceleration; $\hat{v}$ is set to zero at the beginning of the rollout and increases linearly for the first $0.5 |\hat{v}|$ seconds, encouraging the character to accelerate at $2m/s^2$. The parameters of the reward functions used in all the experiments are listed in Table \ref{table:hyper_param}. We demonstrate the environment-centered curriculum learning for all the examples and selectively use the learner-centered curriculum learning for comparison. The resulting motions can be seen in the supplementary video.

We set the starting point of the curriculum $\vc{x}_0$ to $(k_p,k_d)=(2000,2000)$ in all examples. Note that $k_p$ and $k_d$ are the proportional gains and damping gains in two independent SPD controllers that provide balancing and propelling forces respectively. The damping gain used to compute the balancing force is $0.1k_p$ and the proportional gain used to compute the propelling force is $0$.

We use Pydart \cite{pydart}, a python binding of the DART library \cite{DART} to perform multi-body simulation. We simulate the characters at $500$ Hz, and query the control policy every $15$ simulation steps, yielding a control frequency of $33$ Hz. We use the PPO algorithm implemented in the OpenAI Baselines library \cite{baselines} for training the control policies. The control policies used in all examples are represented by feed-forward neural networks with three fully-connected hidden layers, and each hidden layer consists of $64$ units. We fix the sample number to be $20,000$ steps per iteration for all examples. The number of iteration required to obtain a successful locomotion controller depends on the complexity of the task, ranging from $500$ to $1500$, yielding a total sample number between $10$ and $30$ millions. We perform all training using 8 parallel threads on an Amazon EC2 node with 16 virtual cores and 32G memory. Each training iteration takes $25-45$s depending on the degrees of freedoms of the character model, leading to a total training time between $4$ and $15$ hours.

\begin{table*}[ht]
\begin{center}
    \caption{Task and reward parameters}
\label{table:hyper_param}
\begin{tabular}{|l|c|c|c|c|c|c|c|c|}
\hline
Character     &  $\hat{v}$  & $w_v$  & $w_{u_x}$ & $w_{u_y}$ &$w_{u_z}$  &  $w_l$ & $E_a$ & $w_e$  \\
\hline
Simplified Biped &$0$ to $1m/s$  & 3  & 1 & 1 & 1 & 3 & 4  & 0.4 \\
\hline
Simplified Biped& $0$ to $5m/s$  & 3  & 1 & 1 & 1 & 3 &  7  & 0.3 \\
\hline
Quadruped &$0$ to $2m/s$   & 4  & 0.5 & 0.5 & 1 & 3 &  4 & 0.2  \\
\hline
Quadruped &$0$ to $7m/s$  & 4  & 0.5 & 0.5 & 1 & 3 &  11  & 0.35 \\
\hline
Hexapod &$0$ to $2m/s$  & 3  & 1 & 1 & 1 & 3 & 4 & 0.2 \\
\hline
Hexapod &$0$ to $4m/s$  & 3  & 1 & 1 & 1 & 3 &  7 & 0.2  \\
\hline
Humanoid &$0$ to $1.5m/s$   & 3  & 1 &1.5&1& 3 &  6 & 0.3 \\
\hline
Humanoid& $0$ to $5m/s$  & 3  & 1 & 1.5 &1 & 3 &  9 & 0.15  \\
\hline
Humanoid& $0$ to $-1.5m/s$  & 3  & 1 & 1.5 & 1& 3 &  6 & 0.3  \\
\hline
\end{tabular}
\end{center}
\end{table*}

\subsection{Locomotion of Different Morphologies}

\paragraph{Simplified biped} 

\begin{figure}[ht]
  \centering
  \includegraphics[width=\linewidth]{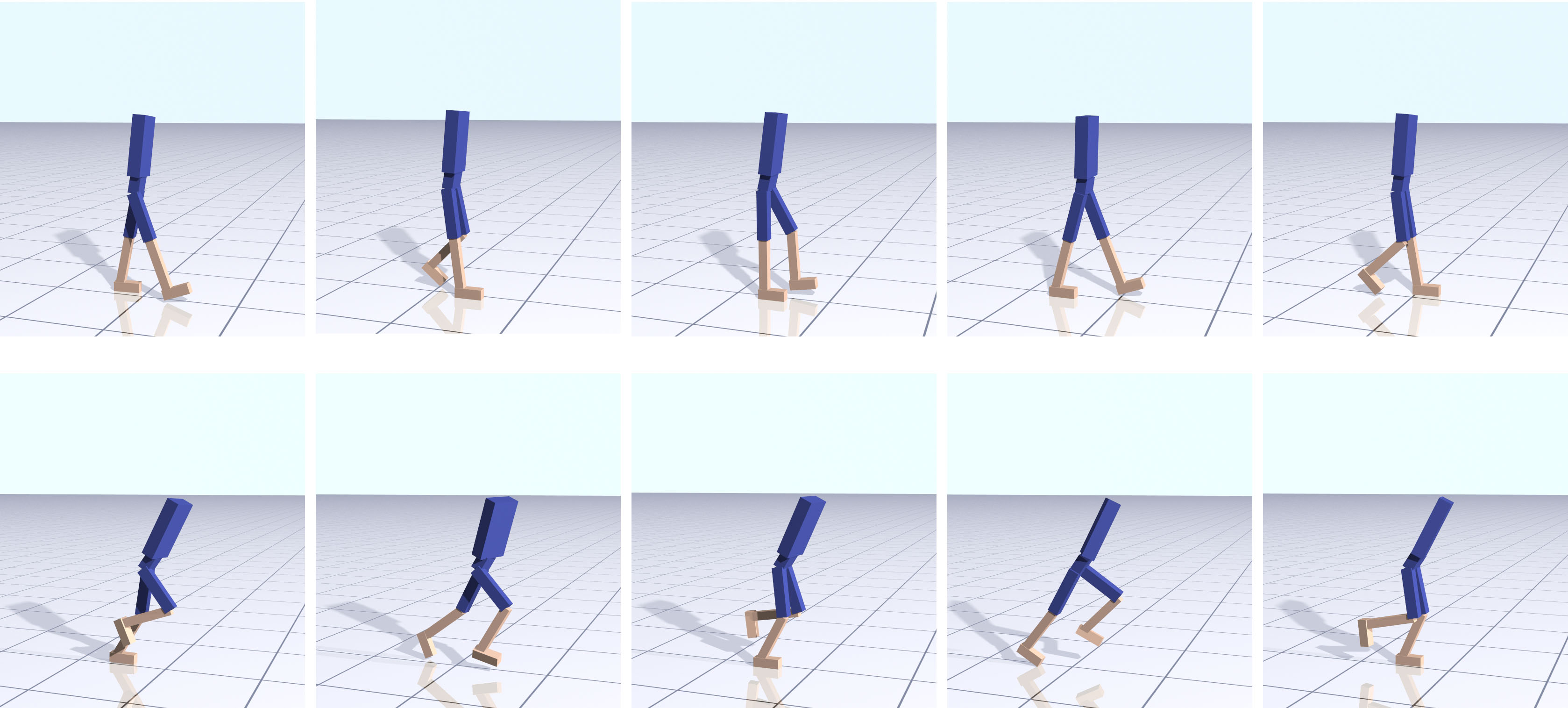}
  \caption{Simplified biped walking (top) and running (bottom). Results are trained using environment-centered curriculum learning and mirror symmetry loss.}
  \label{fig:walker_results}
\end{figure}

Bipedal locomotion has been extensively studied in the literature, and it is a familiar form of locomotion to everyone. Thus we start with training a simplified biped character to perform walking and running. The character has $9$ links and $21$ DOFs, with $1.65$m in height and weighs in total $50$kg. The results can be seen in Figure \ref{fig:walker_results}. As expected, when trained with a low target velocity ($1m/s$), the character exhibits a walking gait. When trained with a high target velocity ($5m/s$), the character uses a running gait indicated by the emergence of a flight phase.

\paragraph{Quadruped}

\begin{figure}[ht]
  \centering
  \includegraphics[width=\linewidth]{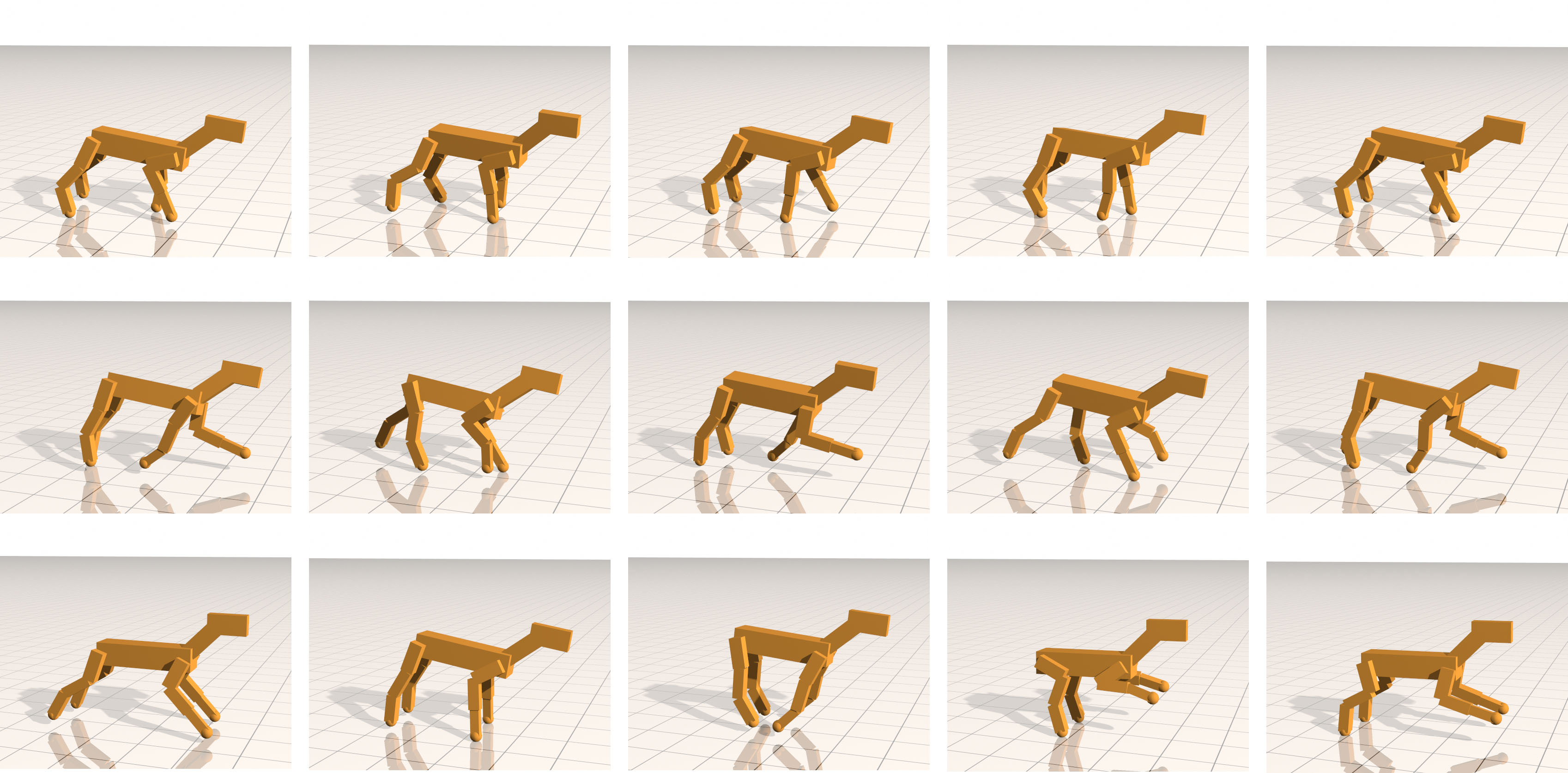}
  \caption{Dog trotting (top, middle) and galloping (bottom). Results are trained using environment-centered curriculum learning and mirror symmetry loss.}
  \label{fig:dog_results}
\end{figure}

Quadrupeds exhibit a large variety of locomotion gaits, such as pacing, trotting, cantering, and galloping. We applied our approach to a quadruped model as shown in Figure\ref{fig:teaser}(b). The model has $13$ links and $22$ DOFs, with a height of $1.15$m and weight of $88.35$kg. As quadrupeds can typically move faster than biped, we trained the quadruped to move at $2$m/s and $7$m/s. The results are shown in Figure \ref{fig:dog_results}. The trained policy results in a trotting gait for low target velocity consistently. For high target velocities, the character learns either trotting or galloping, depending on the initial random seed of the policy.

\paragraph{Hexapod}

\begin{figure}[ht]
  \centering
  \includegraphics[width=\linewidth]{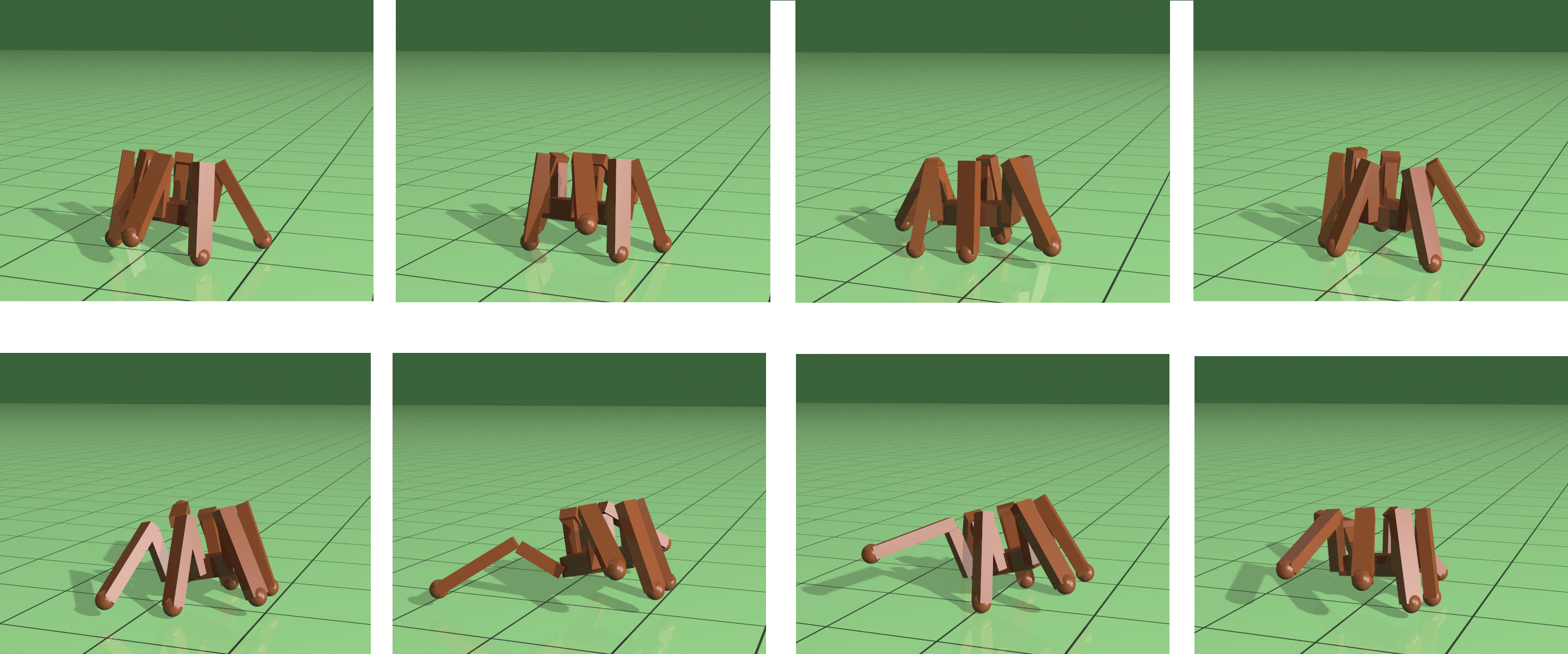}
  \vspace{-3mm}
  \caption{Hexapod moving at $2$m/s (top) and $4$m/s (bottom). Results are trained using environment-centered curriculum learning and mirror symmetry loss.}
  \label{fig:hexapod_results}
\end{figure}

We designed a hexapod creature that has $13$ links and $24$ DOFs, inspired by the body plan of an insect. We trained the hexapod model to move at $2$m/s and $4$m/s. As shown in Figure \ref{fig:hexapod_results}, the hexapod learns to use all six legs to move forward at low velocity, while it lifts the front legs and use the middle and hind legs to 'run' forward at higher velocity.

\begin{figure}[ht]
  \centering
  \includegraphics[width=\linewidth]{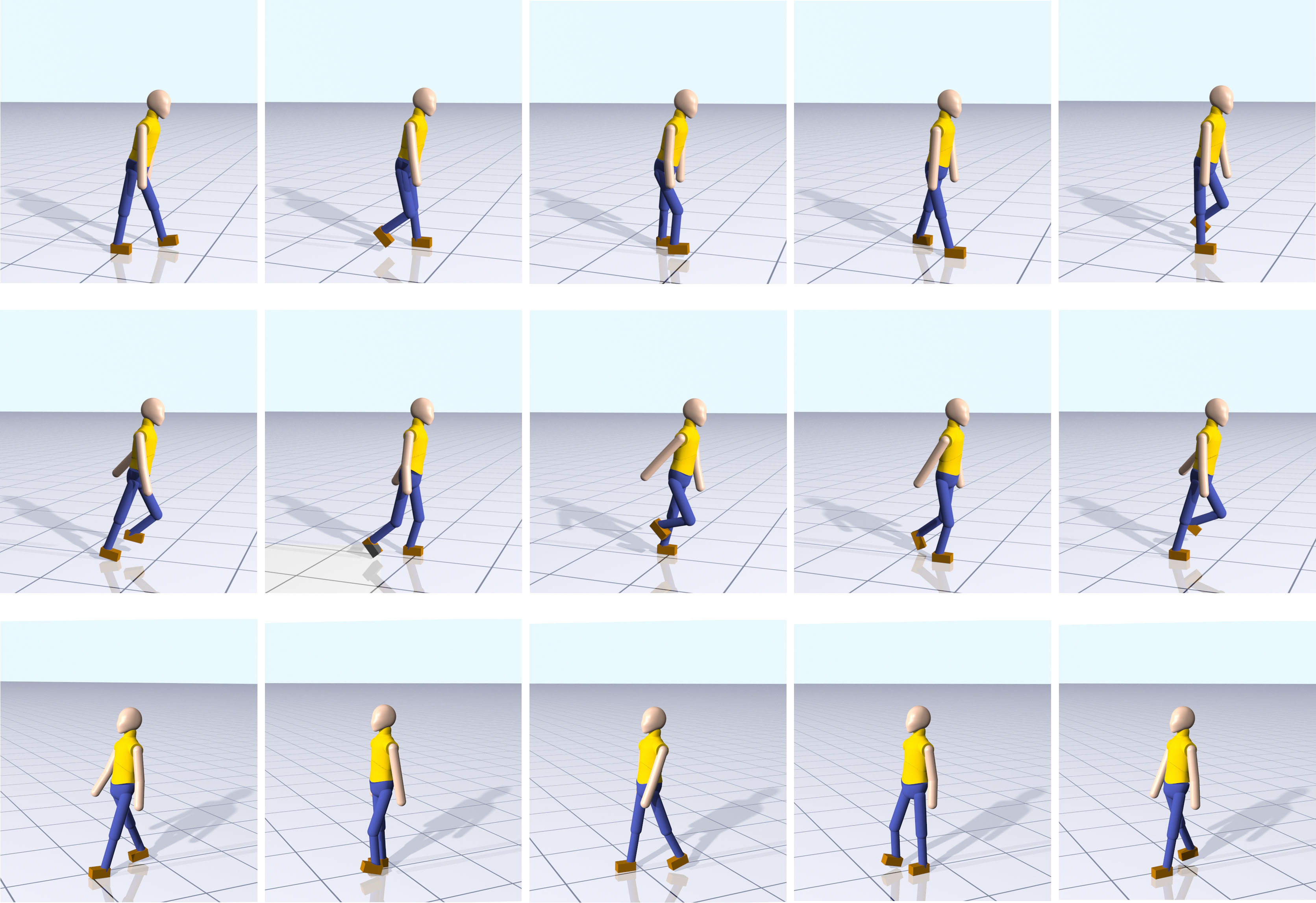}
  \caption{Humanoid walking (top), running (middle) and backward walking (bottom). Results are trained using environment-centered curriculum learning and mirror symmetry loss.}
  \label{fig:human_results}
\end{figure}

\paragraph{Humanoid}
Finally, we trained a locomotion policy for a full humanoid character with a detailed upper body. The character has $13$ links and $29$ DOFs with $1.75$m in height and $76.6$kg in weight. We trained the humanoid model to walk at $1.5$m/s and run at $5$m/s. In addition, we trained the model to walk backward at $-1.5$m/s. We kept the same reward function parameters between forward and backward walking. Results of the humanoid locomotion can be seen in Figure \ref{fig:human_results}. During walking forward and backward, the character learns to mostly relax its arms without much swinging motion. For running, the character learn to actively swing its arms in order to counteract the angular momentum generated by the leg movements, which stabilizes the torso movements during running.

\subsection{Comparison between Learner-centered and Environment-centered Curriculum Learning}

\begin{figure}[ht]
  \centering
  \includegraphics[width=\linewidth]{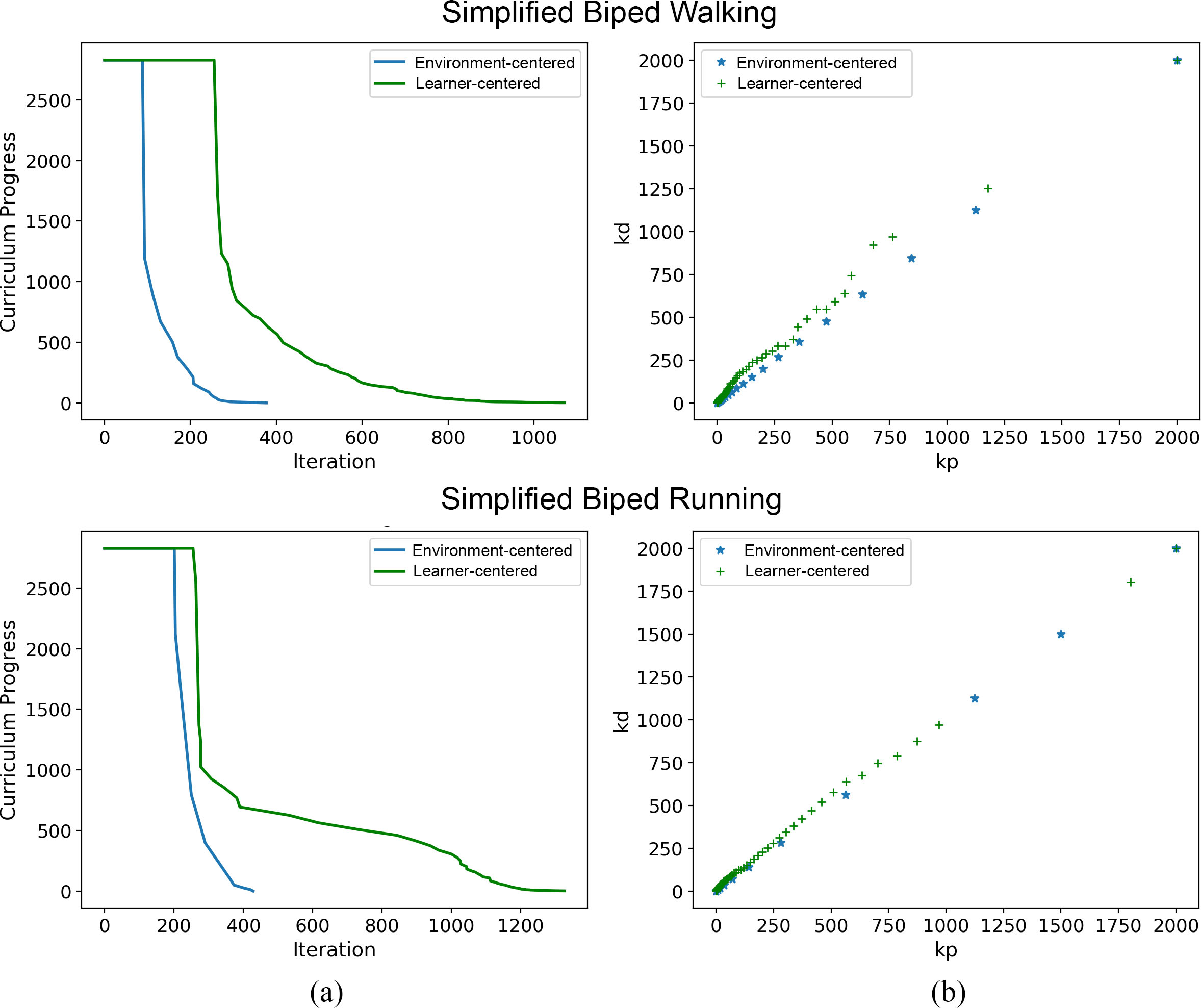}
  \caption{Comparison between environment-centered and learner-centered curriculum learning for simplified biped tasks. (a) Curriculum progress over iteration numbers. $0$ in the $y$ axis means no assistance is provided. (b) Points in the curriculum space visited by the two curriculum update schemes.}
  \label{fig:cl_compare}
\end{figure}

We compare the learner-centered and environment-centered curriculum learning algorithms on the simplified biped model. As demonstrated in the supplementary video, both methods can successfully train the character to walk and run at target velocities with symmetric gaits. We further analyze the performance of the two algorithms by comparing how they progress in the curriculum space, as shown in Figure \ref{fig:cl_compare}. We measure the progress of the curriculum learning with the $l2$ norm of the curriculum parameter $\vc{x}$, since the goal is to reach $0$ as fast as possible. We can see that environment-centered curriculum learning shows superior data-efficiency by generating a successful policy with about half the data that is required for the learner-centered curriculum learning. 

\begin{figure*}[ht]
  \centering
  \includegraphics[width=\linewidth]{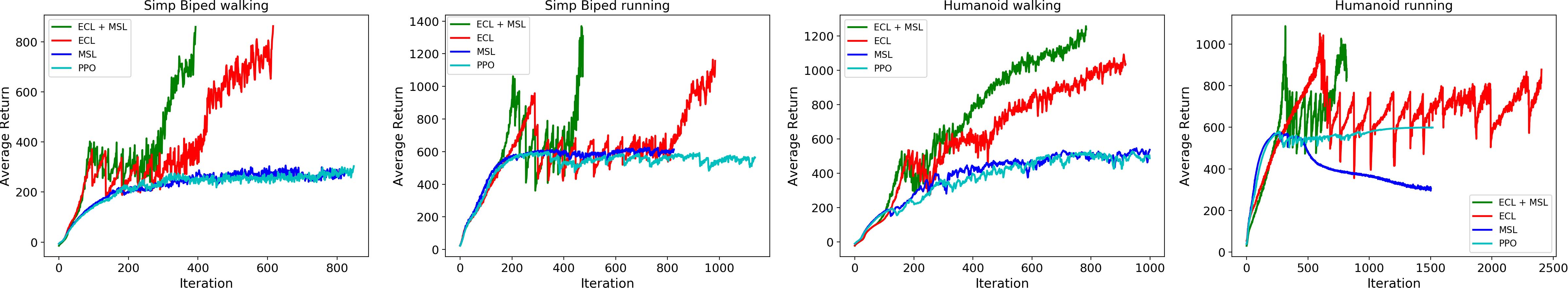}
  \vspace{-3mm}
  \caption{Learning curves for the proposed algorithm and the baseline methods.}
  \label{fig:learning_curves}
\end{figure*}

\subsection{Comparison with Baseline Methods}

\begin{figure}[ht]
  \centering
  \includegraphics[width=\linewidth]{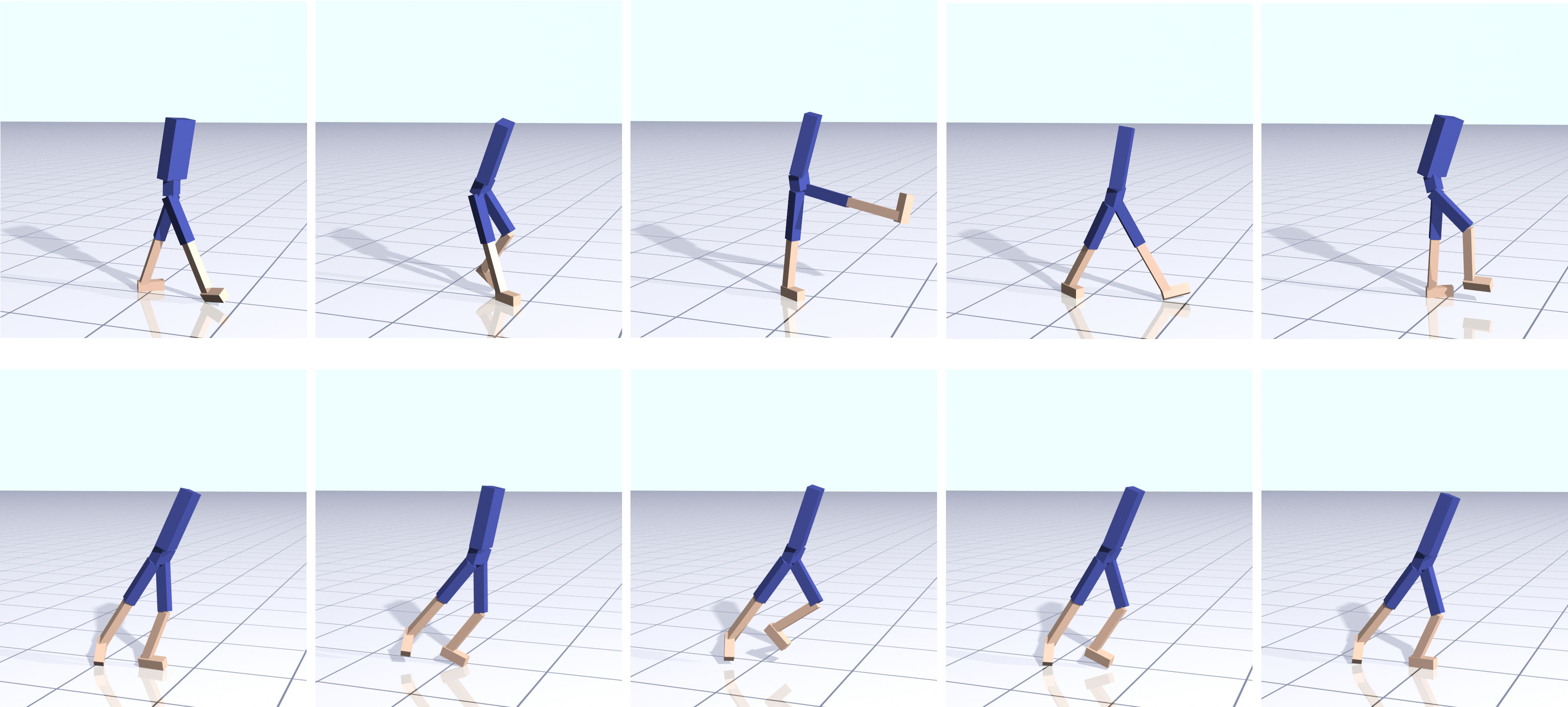}
  \caption{Simplified biped walking (top) and running (bottom). Results are trained using environment-centered curriculum learning only (no mirror symmetry loss).}
    \vspace{-0.4cm}
    \label{fig:walker_asym_results}
\end{figure}

\begin{figure}[ht]
  \centering
  \includegraphics[width=\linewidth]{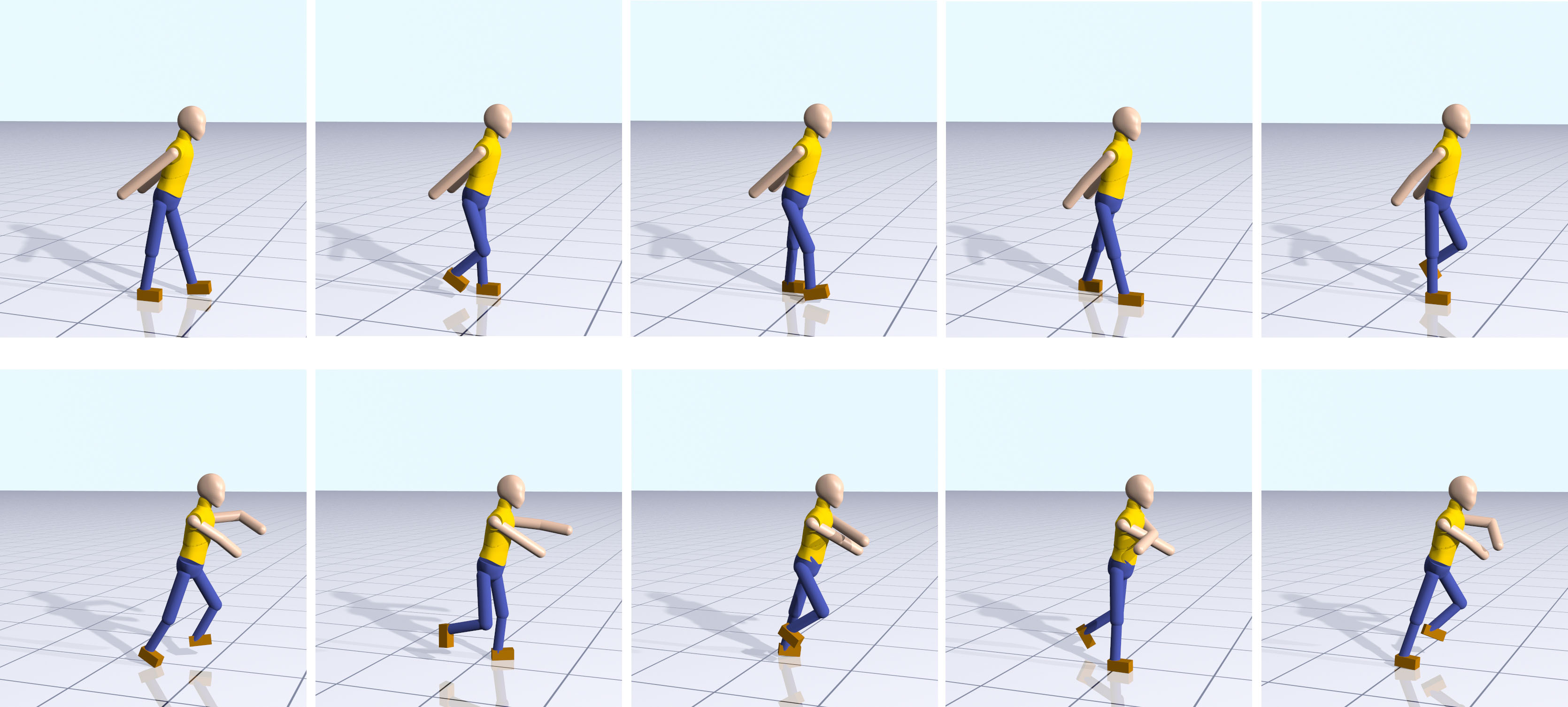}
  \caption{Humanoid walking (top) and running (bottom). Results are trained using environment-centered curriculum learning only (no mirror symmetry loss).}
    \vspace{-0.4cm}
  \label{fig:human_asym_results}
\end{figure}

To demonstrate the effect of curriculum learning and mirror symmetry loss, we compare our method with environment-centered curriculum learning and mirror symmetry loss (ECL + MSL) to three baseline methods: with environment-based curriculum learning only (ECL), with mirror symmetry loss only (MSL) and using vanilla PPO (PPO) with no mirror symmetry loss nor curriculum learning. The baseline methods are trained on the simplified biped character and the humanoid character for both walking and running. The learning curves for all the tests can be seen in Figure \ref{fig:learning_curves}. In all four of these tasks, our approach learns faster than all of the baseline methods. Without curriculum learning (\ie blue and cyan curves), the algorithm typically learns to either fall slowly or stand still (as is shown in the supplementary video). On the other hand, without mirror symmetry loss, the resulting policy usually exhibits asymmetric gaits and the training process is notably slower, which is mostly evident in the running tasks, as shown in Figure \ref{fig:walker_asym_results} and Figure \ref{fig:human_asym_results}.

In addition to the three baseline methods described above, we also trained on the simplified biped walking task using vanilla PPO with a modified reward function, where $w_e$ is reduced to $0.1$. This allows the character to use higher torques with little penalty ("PPO high torque" in Table \ref{table:policy_performance}). While the character is able to walk forward without falling, the motion appears jerky and uses significantly more torque than our results (see supplementary video).

%In addition to the three baseline methods described above, we also trained on the simplified biped walking task using PPO with a modified reward function, where $w_e$ is reduced to $0.1$. This allows the character to use higher torques without being penalized more, thus we name it \textit{PPO high torque}. We were able to train a policy for this one task using PPO with the modified reward function. However, the motion is more jerky and uses significantly more torque than our results (see supplementary video and Table \ref{table:policy_performance}). We have tested this reward modification and PPO on all the other tasks, but we were not able to obtain a successfully trained policy.

To compare the policies quantitatively, we report the average actuation magnitude i.e. $E_e$ and use the Symmetry Index metric proposed by Nigg \etal to measure the symmetry of the motion \cite{nigg1987use}:

\[
SI(X_L, X_R) = 2\frac{|X_L - X_R|}{X_L+X_R} \%,
\]
where $X_L$ and $X_R$ are the average of joint torques produced by the left and right leg respectively. The smaller the value $SI$ is, the more symmetric the gait is. The results can be seen in Table \ref{table:policy_performance}. As expected, policies trained with our method uses less joint torque and produces more symmetric gaits.

%\begin{figure}[ht]
%  \centering
%  \includegraphics[width=6cm]{images/avg_action}
%  \caption{Learning curves for the proposed algorithm and the baseline methods.}
%  \label{fig:avg_action}
%\end{figure}

\begin{table}[ht]
\begin{center}
    \caption{Comparison of trained policies in action magnitude and symmetry. ECL denotes using environment-centered curriculum learning, MSL means mirror symmetry loss and PPO means training with no curriculum learning or mirror symmetry loss. We present results for the successfully trained policies.}
\label{table:policy_performance}
\begin{tabular}{|l|c|c|c|}
\hline
Task     &  Training Setup  & $E_e$ & SI \\
\hline
Simplified Biped walk & ECL+MSL & 2.01 &0.0153\\
\hline
Simplified Biped walk & ECL & 2.98 &0.1126\\
\hline
Simplified Biped walk & PPO high torque & 5.96 &  0.0416 \\
\hline
Simplified Biped run & ECL + MSL &5.57 &0.0026 \\
\hline
Simplified Biped run & ECL &6.052 & 0.4982 \\
\hline
Humanoid walk  & ECL+MSL & 6.2 & 0.0082  \\
\hline
Humanoid walk  & ECL & 7.84 & 0.0685  \\
\hline
Humanoid run  &  ECL+MSL & 17.0976 & 0.0144 \\
\hline
Humanoid run  &  ECL & 18.56 & 0.0391 \\
\hline

\end{tabular}
\end{center}
\end{table}

\subsection{Learning Asymmetric Tasks}

\begin{figure}[ht]
  \centering
  \includegraphics[width=\linewidth]{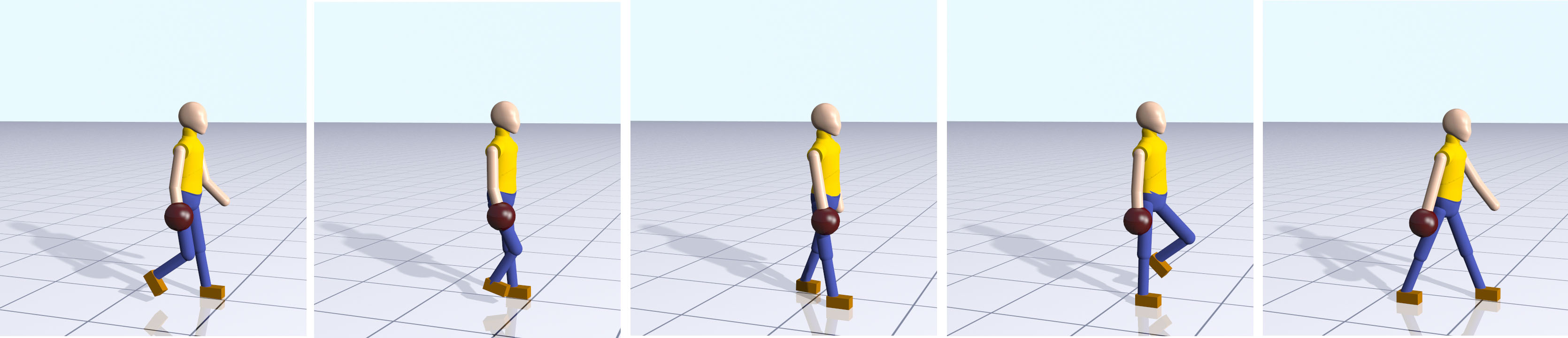}
  \caption{Humanoid walking holding heavy object in right hand. Results are trained using environment-centered curriculum learning and mirror symmetry loss.}
  \vspace{-0.4cm}
  \label{fig:human_heavyobj_result}
\end{figure}

One benefit of encouraging symmetric \emph{actions} rather than symmetric \emph{states} is that it allows the motion to appear asymmetric when desired. As shown in Figure \ref{fig:human_heavyobj_result}, we trained a humanoid that is walking while holding a heavy object ($10$kg) in the right hand. The character uses an asymmetric gait that moves more vigorously on the left side to compensate for the heavy object on the right side. If we chose to enforce symmetry on the states directly, the character would likely use a large amount of torque on the right side to make the poses appear symmetric.

%\subsection{Different Action Spaces}

%So far all the control policies we have trained outputs the joint torques directly. Alternatively, we can have the policy output a target joint position and track it using standard methods such as PID control. We show one such example in ?, where we use target joint position as the action space for the simplified biped character and train it to walk.
% So far the policies doesn't look very good, unless the next batch works, it maynot be worth including this.

%\subsection{Robustness Study}

\section{Discussion}
\begin{figure}[ht]
  \centering
  \includegraphics[width=\linewidth]{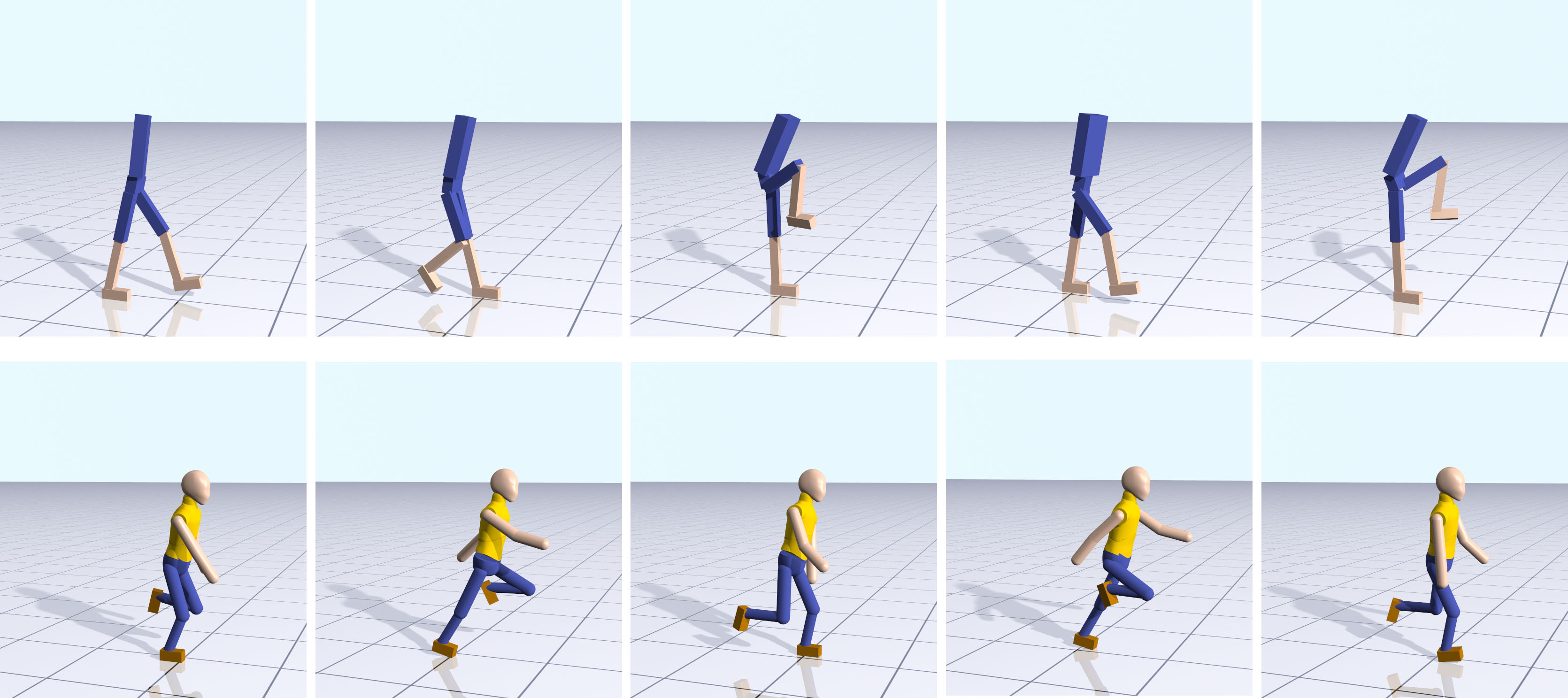}
  \caption{Biped walking with high knees (TOP) and humanoid running with large stride length (BOTTOM) warm-started from our approach.}
  \label{fig:human_stridereward}
  \vspace{-3mm}
\end{figure}

In this work, we intentionally avoid the use of motion examples to investigate whether learning locomotion from biomechanics principles is a viable approach. In practice, it would be desirable to use our policy as a starting point and improve the motion quality by further training with motion examples or additional reward terms. For example, we took the network of the biped walking policy to warm-start another policy learning session, in which the character learns to walk with knees raising up high (Figure \ref{fig:human_stridereward} TOP). We also warm-started from the humanoid running policy to obtain a controller using a bigger stride length (Figure \ref{fig:human_stridereward} BOTTOM). Because the starting policy is already capable of balancing and moving forward, the refinement learning takes only 200 iterations to train. 

Our experiments show that both learner-centered and environment-centered curricula are effective in training locomotion controllers. The learner-centered curriculum is designed to bootstrap learning from the character's existing skill, but the curriculum might end up taking a long and winding path to reach the origin of the curriculum space. On the other hand, the environment-centered curriculum takes a straight path to the origin, but it presents the character with predetermined lessons and ignores the character's current skill, which might result in an overly aggressive curriculum. However, our scheduling algorithm for the environment-centered curriculum ensures that the lessons in two consecutive learning iterations overlap, reducing the chance of the character being thrown into a completely unfamiliar environment and failing immediately. While the learner-centered curriculum requires less hyper parameter tuning, we prefer the environment-centered curriculum for its data efficiency. The main design choice of the environment-centered curriculum lies in the formula of reduction from $\vc{x}_{begin}$ to $\vc{x}_{end}$. Our arbitrarily designed step function works well in all of our examples, but it may be possible to design a different reduction formula that can lead to an even more data-efficient learning algorithm. 

One of the most encouraging results of this work is the emergence of different gaits for different target velocities. Most previous work obtains different gait patterns through contact planning or constraint enforcement \cite{Coros2010,Ye2010OFC}. In contrast, with different target velocity $\hat{v}$, our characters learn speed-appropriate gaits using nearly identical reward functions (Table \ref{table:hyper_param}), without additional engineering effort.

In addition to the parameters of the reward function and hyper parameters in the network, the style of the motion also depends on the kinematic and dynamic properties of the character. A comprehensive sensitivity analysis is beyond the scope of this work. However, we notice that the resulting motion is particularly sensitive to the range of joint torques that each actuator is allowed to generate. In this paper, the torque range is set heuristically based on the perceived strength of each joint. Our preliminary results show that different locomotion gaits can result when different torque range settings. For example, we observed hopping behavior when we greatly increased the leg strength of the humanoid character.

There are many different ways to design the virtual assistant, and some are more effective than others. For example, we tried to provide lateral balance using two walls, one on each side of the character. The initial lesson presented a narrow passage between the two walls so that the character could not possibly fall. As the curriculum proceeded, we then widened the gap between the walls. The results showed that the character learned to intentionally lean on the wall to decrease the penalty of falling. As the gap widened, the character leaned even more until the character fell because the walls were too far apart. We have also attempted to provide propelling assistance through the use of a virtual treadmill. This experiment was unsuccessful even when learning the initial lesson $\vc{x}_0$. Our speculation is that the treadmill does not provide the necessary assistance for learning how to move forward; the learner still needs to put a significant amount of effort towards matching the speed of moving ground. In other words, it is arguable that learning with assistance ($\vc{x}_0$) is just as difficult to learn as without assistance.

%We have also tried to use a treadmill approach to guide the character in place of the forward assistance force. We found this approach to be less effective than providing assistance forces and many times we were not able to train a locomotion controller successfully. We also tried an alternative method for providing balance assistance, where we created a physical narrow passage next to the character. We found that the agent learns to lean on the wall, and were not able to learn to walk in 3D environment through widening the passage.

%In addition to lateral balance and propelling forward, assistance in the other aspects of locomotion may also improve learning performance. For example, assistance torques can be applied to help the character in learning to turn.
%During our experiments, we have tried providing assistive forces from below to prevent character from falling, but the character learns to jump forward, lift both feet and fly.

%For most of our experiments, we were not able to obtain successful policy trained using PPO alone. On the other hand, researchers have obtained working policies for high-dimensional control policies with PPO though the motions are less natural \cite{heess2017emergence}. The discrepancy maybe due to the design difference in the characters, reward functions and hyper-parameter settings. It would be interesting to see how our approach works in those examples.

\section{CONCLUSION}

We have demonstrated a reinforcement learning approach for creating low-energy, symmetric, and speed-appropriate locomotion gaits. One element of this approach is to provide virtual assistance to help the character learn to balance and to reach a target speed. The second element is to encourage symmetric behavior through the use of an additional loss term.  When used together, these two techniques provide a method of automatically creating locomotion controllers for arbitrary character body plans. We tested our method on the lower half of a biped, a full humanoid, a quadruped, an a hexapod and demonstrated learning of locomotion gaits that resemble the natural motions of humans and animals, without prior knowledge about the motion. Because our method generalizes to other body plans, an animator can create locomotion controllers for characters and creatures for which there is no existing motion data.

Although our characters demonstrate more natural locomotion gaits comparing to existing work in DRL, the quality of the motion is still not on a par with previous work in computer animation that exploits real-world data. Further investigation on how to incorporate motion capture data,  biological-based modeling, and policy refinement (see Figure \ref{fig:human_stridereward}) is needed. In addition, our work is only evaluated on terrestrial locomotion with characters represented by articulated rigid bodies. One possible future direction is to apply the curriculum learning to other types of locomotion, such as swimming, flying, or soft-body locomotion.

\begin{acks}
We thank Charles C. Kemp, Jie Tan, Ariel Kapusta, Alex Clegg, Zackory Erickson and Henry M. Clever for the helpful discussions. This work was supported by NSF award IIS-1514258 and AWS Cloud Credits for Research.
\end{acks}

\bibliographystyle{ACM-Reference-Format}
\bibliography{aaatemplate}
\end{document}